# Texture Based Image Segmentation of Chili Pepper X-Ray Images Using Gabor Filter

M. Rajalakshmi

Research Scholar,

Dr. P. Subashini

Professor,

Department of Computer Science, Avinashilingam Institute for Home Science and *Higher Education for Women, Coimbatore, India*

**Abstract: Texture segmentation is the process of partitioning an image into regions with different textures containing a similar group of pixels. Detecting the discontinuity of the filter's output and their statistical properties help in segmenting and classifying a given image with different texture regions. In this proposed paper, chili x-ray image texture segmentation is performed by using Gabor filter. The texture segmented result obtained from Gabor filter fed into three texture filters, namely Entropy, Standard Deviation and Range filter. After performing texture analysis, features can be extracted by using Statistical methods. In this paper Gray Level Co-occurrence Matrices and First order statistics are used as feature extraction methods. Features extracted from statistical methods are given to Support Vector Machine (SVM) classifier. Using this methodology, it is found that texture segmentation is followed by the Gray Level Co-occurrence Matrix feature extraction method gives a higher accuracy rate of 84% when compared with First order feature extraction method.**

## I. INTRODUCTION

Image segmentation is a fundamental process in many images, video and computer vision applications. It decomposes an image into several basic components, which preferably correspond into real-world objects. Region-based image segmentation is a popular approach to image segmentation. In this approach, an image is partitioned into connected regions by grouping neighboring pixels of similar features, and adjacent regions are then merged under some criterion such as homogeneity of features in neighboring regions. Features of interest often include color, texture, shape, etc. Texture is one common feature used in image segmentation.

Texture segmentation has long been an important chore in the image processing field. Mainly, it aims at segmenting a textured image into several regions having similar patterns. An effective and efficient texture segmentation method will be very useful in applications like analysis of aerial images, biomedical images, quality control and seismic images as well as automation of industrial applications. Like other segmentation problems, segmentation of textures requires choice of proper texture-specific features with good discriminative power [5]. Texture segmentation is a tricky problem because one usually does not know the advances what type of textures exist in an image, how many unusual textures there, and what regions in the image have which textures [6].

Texture analysis refers to the characterization of regions in an image by their texture content. Texture analysis attempt to measure intuitive qualities described by texture terms such as roughness, smoothness, silky, or bumpiness as a function of the spatial variation in pixel intensities. In this logic, the roughness or bumpiness refers to variations in intensity values, or gray levels. The analysis of texture requires identification of those texture attributes which can be used for segmentation, classification, pattern recognition, or shape computation, and the development of computational approaches for accomplishing these tasks.

Food safety and demand for high quality food is a great concern today. Generally chili pepper has been contaminated by fungus or toxin during harvesting, transportation. The chilies were collected from the village and get contaminated fungus. To ensure its quality, x-ray images of chilies are taken. The chili x-ray images are preprocessed and then applied into segmentation process.

In this paper, pre-processed chili x-ray images were taken as input. Range Filtering, Entropy filtering and standard deviation filtering along with Gabor filter are used to segment chili images. The 2-D Gabor filters have been shown here particularly for analyzing textured images containing highly specific frequency or orientation characteristics [8]. Gabor filters have been applied by Clark et ul. [9] to both natural and artificial textures, and by Turner [10] to artificial textures similar to those often used in psychophysical experiments [1]. The images are used in this paper are given in fig 1.

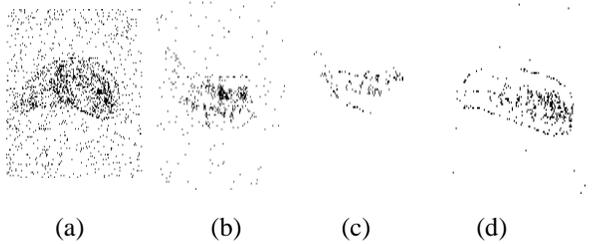

(a)      (b)      (c)      (d)

Figure 1: Pre-processed Chili X-ray images (a)-(d)

The paper is organized as follows: section 2 describes the methodology. In section 3, the results obtained with proposed methods are shown. Finally, section 4 contains some concluding remarks and all the references being made for completion of this work.

## II.  METHODOLOGY

Texture is an important characteristic for the analysis of many kinds of images and also has an essential role in many computer vision and image processing applications by including rich source of visual information. Texture regions are extracted from a chili image by using Gabor filter method. The resultant chili image obtained from Gabor texture segmentation has a variety of sub regions of textures. The segmented images are given as input to Range filters, Standard Deviation filtering, and Entropy filtering methods. Different texture regions extracted from this are given as the input to texture feature extraction methods such as First order, GLCM and Geometric methods. Extracted features are analyzed by using Support Vector Machine classifier. The proposed work is shown in fig 2.

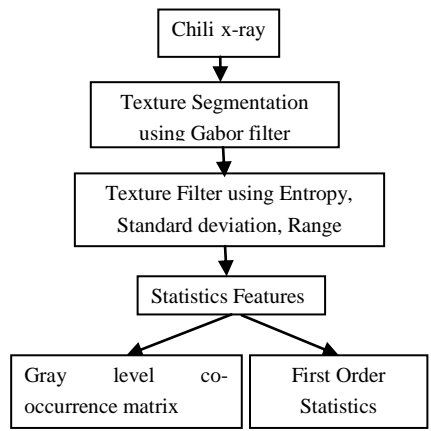

Fig.2 Flow of Proposed Work

### A.  Gabor Filter

Gabor filter is a linear filter whose impulse response is defined by a harmonic function multiplied by a Gaussian function. The nature of good spatial and spatial frequency localization of Gabor filters makes them extensively used for texture segmentation. Gabor filter related segmentation paradigm is based on the filter bank model in which several filters are applied simultaneously to an input image. The filter's center on a particular range of frequencies. When an input image contains different texture areas, then it will provide more filter output sub images. The local frequency differences between areas will provide output images. Each Gabor filter is a precise one by Gabor Elementary function (GEF). GEFs can perform joint space decomposition. Gabor filters are widely used for texture segmentation because of their good spatial and spatial-frequency localization. Gabor Elementary Functions were first defined by Gabor and later extended to 2-D by Daugman. A GEF is derived from the following:

$$h(s,t) = g(s^{'}, t^{'})\exp\left[j2\pi(M_s + N_t)\right] \quad (1)$$

where $(s^{'}, t^{'}) = (s\cos\theta + t\sin\theta, -s\sin\theta + t\cos\theta)$ represent rotated spatial-domain rectilinear coordinates. Let (m, n) denotes frequency-domain rectilinear coordinates, (M, N) represents particular 2-D frequency. The complex exponential is a 2-D complex sinusoid at a frequency $F = \sqrt{M^2 + N^2}$ and $\varphi = \tan^{-1} N/M$ specifies the orientation of the sinusoid [2]. The function g (x, y) is the 2-D Gaussian

$$g(s,t) = \frac{1}{2\pi\sigma_s\sigma_t}\exp\left\{-\frac{1}{2}\left[\left(\frac{s}{\sigma^s}\right)^2 + \left(\frac{t}{\sigma^t}\right)^2\right]\right\} \quad (2)$$

where, $\sigma_s$ and $\sigma_t$ denotes the spatial extent and bandwidth of the filter [16]. GEF is a Gaussian that is modulated by a complex sinusoid. It can be shown that the Fourier transform of h(s, t) is given by

$$H(m,n) = \exp\left\{-\frac{1}{2}\left[\left(\sigma_s[m-M]\right)^2 + \left(\sigma_t[n-N]\right)^2\right]\right\} \quad (3)$$

where $[(m-M)', (n-N)'] = [(m-M)\cos\theta + (n-N)\sin\theta, -(m-M)\sin\theta + (n-N)\cos\theta]$.

Thus, from equation 3, the GEF's frequency response has the shape of a Gaussian. The major and minor axis widths of Gaussian's are determined by $\sigma_s$, $\sigma_t$, and it is rotated by an angle θ with respect to the positive m-axis and it is centered about the frequency (M, N). Thus, GEF acts as a band-pass filter. In most cases, letting $\sigma_s = \sigma_t = \sigma$ is a reasonable design choice [15]. It is assumed that $\sigma_s = \sigma_t = \sigma$, then the parameter $\theta$ is no more needed and the equation of GEF is reduced into

$$h(s,t) = \frac{1}{2\pi\sigma^2} \exp\left\{-\frac{s^2+t^2}{2\sigma^2}\right\} \exp\left[j2\pi(M_s+N_t)\right] \quad (4)$$

Gabor Filter $O_h$ is defined by

$$O(s,t) = O_h(i(s,t)) = |\, i(s,t) \otimes h(s,t)\,| \quad (5)$$

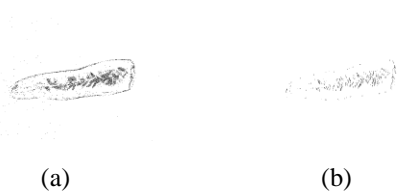

(a)                              (b)

Fig 3. Texture result Image of Gabor filter (a) Pre-processed image (b) Gabor Filtered image

where i is the input image and O is the output image. The results of the texture segmentation of chili images are shown in fig 3.

### B.  Texture  Filtering

In this work, texture filtering is performed by Range filtering, Standard Deviation filtering and Entropy Filtering. Range filtering calculates the local range of an input image, whereas standard deviation filtering calculates the local standard deviation of an input image. If it is a range filtering, each pixel in the output image is the range, i.e., the variation between the highest and the lowest pixels, of the consequent input pixel and the pixels adjacent it in a chili image. The standard deviation filter does similar operation, but it calculates the standard deviation of the pixels surrounding the corresponding pixel in the input chili image. The Entropy filter calculates the entropy value of the 9-by-9 neighborhood around corresponding pixel in the input chili image. The resultant segmented Images are shown in Table 1.

Table 1. Result images of Texture filters

| S.no | Gabor filtered Image | Range filtered Image | Standard Deviation filtered image | Entropy filtered image |
|---|---|---|---|---|
| 1 |  |  |  |  |
| 2 |  |  |  |  |
| 3 |  |  |  |  |
| 4 |  |  |  |  |

Texture analyzing requires calculations of various features of textures. In this work, statistical texture features were extracted from segmented images. GLCM method is used in analyzing the spatial distributions and relationships between the gray levels of an image. Textures are characterized according to statistical measures calculated from the intensity values of pixels. By using statistical operators to those pixels, texture feature descriptors are calculated. Texture feature descriptors can be classified into two groups: first-order texture features and GLCM texture features [12]. Finally, the Features are analyzed by using SVM classifier.

### C. First Order Texture Features

First-Order statistics measure the properties of an individual pixel values. First order statistics are limited as texture descriptors because they carry no information about comparative position of pixels with respect to one another. First order statistical feature measures the chance of observing gray value intensities at randomly chosen location in the given input chili image [13].

### D. Gray Level Co-Occurrence Matrices

The popular second order texture method which is used for feature extraction is GLCM. Generally, the second- and higher-order statistics measures the properties of two or more pixel values occurring at specific locations relative to each other [6]. The relationships between neighboring pixels are changed into matrix called Gray Level Co-occurrence Matrix (GLCM). GLCMs are constructed to record the relative frequencies of each pixel pair $Pij$ with different gray levels $i$ and $j$ that are separated by a specified neighborhood distance $d$ and direction $\theta$. The most commonly used directions are $\theta = 0°, 45°, 90°, 135°$ [12]. In this work thirteen texture feature is extracted by using GLCM method.

### III. IMPLEMENTATION AND EXPERIMENTAL RESULTS

First order textures of each chili image were examined through six metrics. They are mean, skewness, kurtosis, standard deviation, entropy, and smoothness. The results of the first order texture method of a chili pepper image are shown in Table 2, 3, 4. The abbreviations are used as: ME : Mean, SKEW : Skewness, KUR: Kurtosis, STD : Standard Deviation, ENT : Entropy, STH : Smoothness.

Table2. First Order Statistics of Range-filtered image

| S.no | ME | SKEW | KUR | STD | ENT | STH |
|------|------|------|------|------|------|------|
| 1 | 22.107 | 2.943 | 9.675 | 70.873 | 0.552 | 1.000 |
| 2 | 8.0914 | 5.3486 | 29.638 | 44.150 | 0.276 | 1.000 |
| 3 | 2.953 | 9.138 | 84.593 | 26.913 | 0.128 | 0.999 |
| 4 | 8.4 | 5.238 | 28.461 | 44.995 | 0.276 | 1.000 |

Table3. First Order Statistics of Standard Deviation filtered image

| S.no | ME | SKEW | KUR | STD | ENT | STH |
|------|------|------|------|------|------|------|
| 1 | 2.555 | 9.910 | 99.392 | 24.825 | 0.144 | 0.998 |
| 2 | 0.877 | 17.078 | 293.186 | 14.599 | 0.061 | 0.995 |
| 3 | 0.000 | 0.062 | 3.904 | 0.004 | 0.000 | 0.001 |
| 4 | 1.807 | 11.812 | 140.760 | 20.965 | 0.104 | 0.998 |

Table4. First Order Statistics of Entropy filtered image

| S.no | ME | SKEW | KUR | STD | ENT | STH |
|------|------|------|------|------|------|------|
| 1 | 7.356 | 5.645 | 32.911 | 42.075 | 0.267 | 0.999 |
| 2 | 4.060 | 7.756 | 61.229 | 31.446 | 0.172 | 0.999 |
| 3 | 2.183 | 10.707 | 115.804 | 23.094 | 0.111 | 0.998 |
| 4 | 4.804 | 7.089 | 51.311 | 34.175 | 0.189 | 0.999 |

The features extracted from Gray level co-occurrence matrix are shown here in the Table

5,6,7. The abbreviations are used as: ASM: Angular System moment, CON: Contrast, VAR: Variance, SA: Sum, Average, SV: Sum Variance, SE: Sum Entropy, ENT: Entropy, DV: Difference Variance, DE: Difference Entropy, IMC1: Information measure of correlation1, IMC2: Information measure of correlation2 [13].

Table5. GLCM features of Range filtered Image

| ASM | CON | COR | VAR | IDM | SA | SV | SE | ENT | DV | DE | IMC1 | IMC2 |
|------|------|------|------|------|------|------|------|------|------|------|------|------|
| 0.8229 | 0.6186 | 0.9198 | 6.3859 | 0.9944 | 3.2124 | 22.8175 | 0.3818 | 0.3957 | 0.6186 | 0.1066 | -0.7462 | 0.6129 |
| 0.9317 | 0.2344 | 0.9217 | 2.9524 | 0.9979 | 2.4439 | 10.8965 | 0.1762 | 0.1814 | 0.2344 | 0.0466 | -0.7804 | 0.4552 |
| 0.9742 | 0.0962 | 0.9138 | 1.6926 | 0.9991 | 2.1621 | 6.4712 | 0.0795 | 0.0816 | 0.0962 | 0.0221 | -0.7816 | 0.3152 |
| 0.9293 | 0.2373 | 0.9236 | 3.0294 | 0.9979 | 2.4611 | 11.1737 | 0.1813 | 0.1864 | 0.2373 | 0.0472 | -0.7824 | 0.4616 |

Table6. GLCM features of Standard Deviation filtered Image

| ASM | CON | COR | VAR | IDM | SA | SV | SE | ENT | DV | DE | IMC1 | IMC2 |
|------|------|------|------|------|------|------|------|------|------|------|------|------|
| 0.9756 | 0.1353 | 0.8577 | 1.5859 | 0.9988 | 2.1388 | 6.0098 | 0.0788 | 0.082 | 0.1353 | 0.0313 | -0.6709 | 0.2819 |
| 0.9915 | 0.0471 | 0.8567 | 1.1807 | 0.9996 | 2.0476 | 4.6734 | 0.0319 | 0.0329 | 0.0471 | 0.0127 | -0.6875 | 0.1841 |
| 0.9993 | 0.0035 | 0.8611 | 0.9851 | 1 | 2.0036 | 4.0476 | 0.0034 | 0.0034 | 0.0035 | 0.0013 | -0.7052 | 0.0612 |
| 0.9828 | 0.1013 | 0.8505 | 1.407 | 0.9991 | 2.0984 | 5.4156 | 0.0583 | 0.0605 | 0.1013 | 0.0229 | -0.6771 | 0.245 |

Table7. GLCM features of Entropy filtered Image

| ASM | CON | COR | VAR | IDM | SA | SV | SE | ENT | DV | DE | IMC1 | IMC2 |
|------|------|------|------|------|------|------|------|------|------|------|------|------|
| 0.9363 | 0.2509 | 0.9077 | 2.7652 | 0.9977 | 2.4024 | 10.1739 | 0.1694 | 0.1749 | 0.2509 | 0.0506 | -0.7475 | 0.434 |
| 0.964 | 0.1344 | 0.9115 | 1.9596 | 0.9988 | 2.2222 | 7.3826 | 0.1059 | 0.1092 | 0.1344 | 0.0321 | -0.7581 | 0.3533 |
| 0.9802 | 0.0778 | 0.9052 | 1.5004 | 0.9993 | 2.1193 | 5.7856 | 0.0646 | 0.0665 | 0.0778 | 0.0206 | -0.7532 | 0.2778 |
| 0.9579 | 0.1443 | 0.9197 | 2.1446 | 0.9987 | 2.2636 | 8.0443 | 0.1203 | 0.1238 | 0.1443 | 0.0346 | -0.7696 | 0.3788 |

*A. Svm Classifier*

The SVM classifier is widely used as classifier in Bioinformatics because of its high accuracy, dealing with high-dimensional data [4]. SVMs work is based on kernel methods. A kernel method is an algorithm that depends on the data only through dot-products. Kernel function can replace the dot product when the dimensional feature space is not huge [3]. The SVM has the ability to make non-linear decision boundaries using methods intended for linear classifiers. Also, SVM can be applied as a classifier in non clear specified fixed-dimensional vector space representation. While training SVM the important things to be considered there are: how to preprocess the data, what kernel we should use, and setting the parameters of the kernel and SVM.

Linear classifier is worked based on a linear discriminant function which is in the form of:

$$f(x) = w^T x + b \qquad (6)$$

The vector w is to be the weight vector, and b is the bias, $x_i$ will denote the input patterns. The hyperplane defined as

$$x : f(x) = w^T x + b = 0 \qquad (7)$$

Divides the region into either positive side or in the negative side[7].

*B. Performance Rate*

Performance rate is a tool to evaluate the prediction performance of the working models. The performance measures are used in this research are precision,recall and F1. The measures are calculated by using the following equations:

$$Precision = P = \frac{TP}{TP+FP} \qquad (8)$$

$$Recall = R = \frac{TP}{TP+FN} \qquad (9)$$

$$F1 = \frac{TP+TN}{TP+FP+TN+FN} \qquad (10)$$

where TP is the number of positive samples in the actual first class, FP is the number of samples in second class, but predicted to be in first class, FN is the number of samples in first class, but predicted to be in the second class, TN is the number of negative samples in second class[17].

The texture features which are extracted by using first order statistical methods and GLCM are shown in the following table 14. From the table, it infers that first order feature value yields less accuracy in all three filtering methods. The GLCM feature value produces higher accuracy than the first order statistics extraction method. Among three texture filters, range filtering produces higher accuracy than the other two filtering methods. From the result obtained above, it concludes that the Range filtering method with GLCM based feature extraction method can be followed by SVM classification provides higher accuracy than the first order statistics feature extraction method.

Table8. Classification accuracy of Texture filtering methods

| Method | Range filter | | | Standard Deviation filter | | | Entropy filter | | |
|---|---|---|---|---|---|---|---|---|---|
| | Precision | Recall | F1 | Precision | Recall | F1 | Precision | Recall | F1 |
| First Order | 0.61 | 0.80 | 0.72 | 0.53 | 0.70 | 0.64 | 0.53 | 0.58 | 0.56 |
| GLCM | **0.84** | **0.84** | **0.84** | 0.61 | 0.72 | 0.68 | 0.85 | 0.73 | 0.76 |

## IV.     CONCLUSION

The Chili pepper x-ray images are texture segmented using Gabor filter. Each Gabor filtered image is fed into three texture filters, namely Range filter, Standard Deviation filter, and Entropy filter. After filtering, to analyze the internal characteristics, textural features are extracted from each chili x-ray image. In textural feature extraction, first order and Gray Level Co-occurrence Matrix features are extracted from each segmented image. Extracted features are given to SVM Classifier to predict the results. The accuracy of classifying shows that the Range filtering followed of GLCM yields higher accuracy, i.e. 84% than the first order feature extraction method. Finally, at this conclusion says that the Range filtering method along with GLCM based feature extraction with SVM classification provides better results.